\documentclass[lettersize,journal]{IEEEtran}

\usepackage{
amsmath,amsfonts}
\usepackage{algorithmic}
\usepackage{algorithm}
\usepackage{array}
\usepackage{textcomp}
\usepackage{stfloats}
\usepackage{url}
\usepackage{verbatim}
\usepackage{graphicx}
\usepackage{cite}

\usepackage{amsmath,amsfonts,amssymb}
\usepackage{graphicx}
\usepackage{url}
\usepackage{cite}

\usepackage{orcidlink}

\begin{document}

\title{Effects of Wrist-Worn Haptic Feedback\\ on Force Accuracy and Task Speed\\ during a Teleoperated Robotic Surgery Task}

\author{%
    Brian Vuong\,\orcidlink{0000-0002-2936-5260},~\IEEEmembership{Student Member,~IEEE,}
    Josie Davidson,
    Sangheui Cheon\,\orcidlink{0000-0003-4567-8901},~\IEEEmembership{Student Member,~IEEE,}
    
    Kyujin Cho\,\orcidlink{0000-0003-2555-5048},~\IEEEmembership{Member,~IEEE,}
    Allison M. Okamura\,\orcidlink{0000-0002-6912-1666},~\IEEEmembership{Fellow,~IEEE}
}



\thispagestyle{empty}




\maketitle

\begin{abstract}
Previous work has shown that the addition of haptic feedback to the hands can improve awareness of tool-tissue interactions and enhance performance of teleoperated tasks in robot-assisted minimally invasive surgery. However, hand-based haptic feedback occludes direct interaction with the manipulanda of surgeon console in teleoperated surgical robots. We propose relocating haptic feedback to the wrist using a wearable haptic device so that haptic feedback mechanisms do not need to be integrated into the manipulanda. However, it is unknown if such feedback will be effective, given that it is not co-located with the finger movements used for manipulation. To test if relocated haptic feedback improves force application during teleoperated tasks using da Vinci Research Kit (dVRK) surgical robot, participants learned to palpate a phantom tissue to desired forces. A soft pneumatic wrist-worn haptic device with an anchoring system renders tool-tissue interaction forces to the wrist of the user. Participants performed the palpation task with and without wrist-worn haptic feedback and were evaluated for the accuracy of applied forces. Participants demonstrated statistically significant lower force error when wrist-worn haptic feedback was provided. Participants also performed the palpation task with longer movement times when provided wrist-worn haptic feedback, indicating that the haptic feedback may have caused participants to operate at a different point in the speed-accuracy tradeoff curve.
\end{abstract}


\section{Introduction}

Robotic-Assisted Minimally Invasive Surgery (RMIS) enables precise and less invasive surgical interventions by offering surgeons enhanced dexterity, scaled motions, and a magnified 3D view of the surgical field. These systems have been associated with reduced patient recovery times, minimal scarring, and improved surgical outcomes compared to traditional open or laparoscopic techniques. 

\begin{figure}[t]
    \centering
    \begin{minipage}{0.48\textwidth}
        \includegraphics[width=\textwidth]{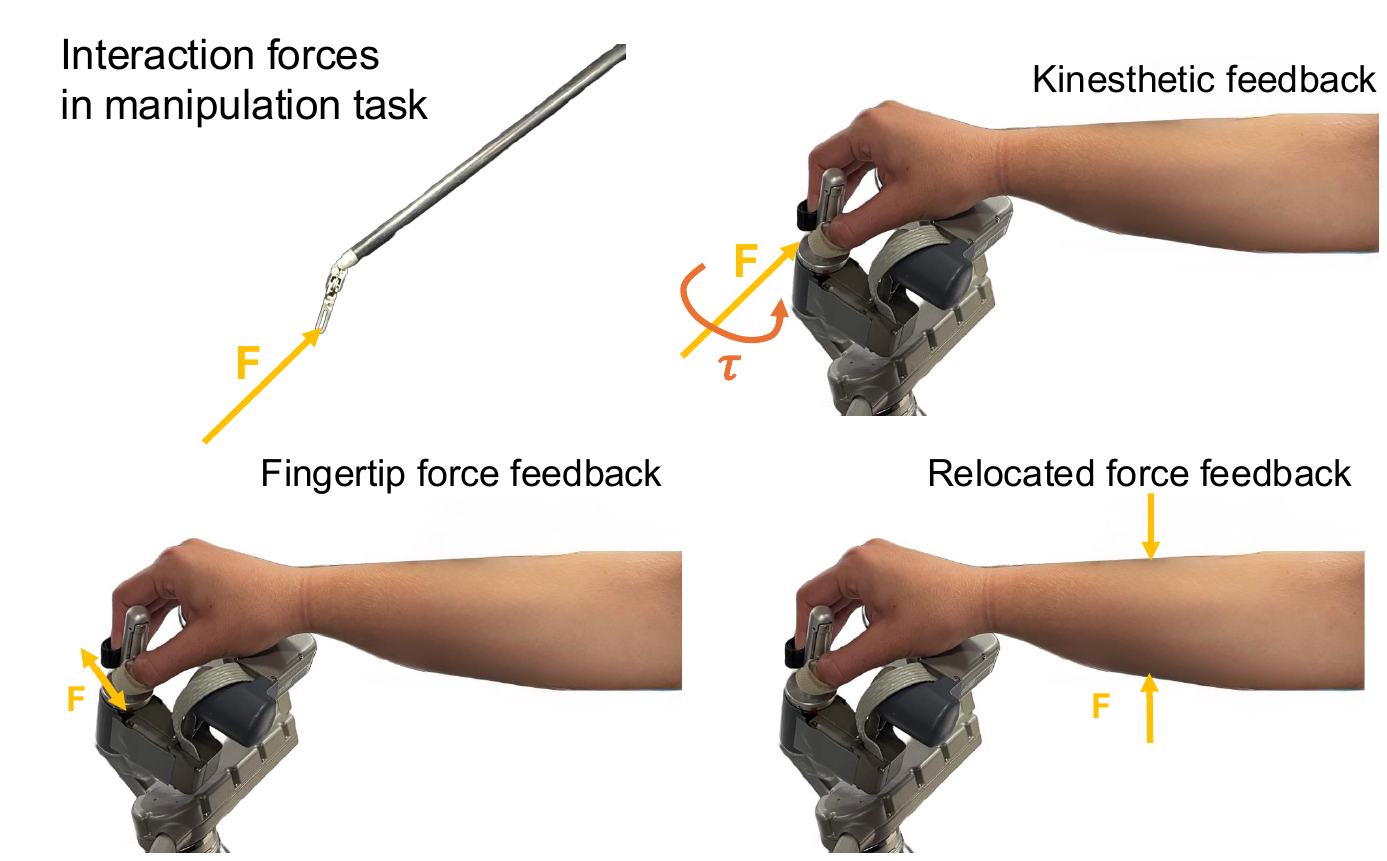}
    \end{minipage}
    \caption{Interaction forces in a teleoperated manipulation tasks can be rendered as either kinesthetic feedback through the manipulandum, cutaneous feedback on the fingertips, or cutaneous feedback relocated to other locations, such as the forearm.}
    \label{fig:feedback_types}
\end{figure}

Haptic feedback, encompassing both kinesthetic and cutaneous sensations, is integral to safe and efficient surgical performance. In traditional open surgery, surgeons rely heavily on their sense of touch to gauge tissue properties and apply appropriate forces. RMIS, however, relies primarily on visual cues to approximate force and stiffness through tissue deformation \cite{kuschel2010combination} \cite{cellini2013visual}. This includes visual cues that indicate tissue damage due to excessive force or blanching of tissue resulting from restricted blood flow \cite{Chua2020}. However, this association between applied force and visual cues is largely heuristic \cite{Klatzky2014}, which can result in suboptimal outcomes due to the inherent limitations of visual cues. These challenges underscore the necessity of developing haptic feedback systems tailored to RMIS, capable of restoring the sense of touch to surgeons without compromising the ergonomics of surgical workflows. Incorporation of integrated force feedback in teleoperated RMIS systems improves surgeon's ability to perceive and modulate interaction forces during tissue manipulation. RMIS teleoperation with haptic feedback significantly reduces the risk of tissue damage and provides a significant benefit to surgical precision, particularly in delicate operations such as blunt dissection \cite{Wagner2007}, tissue grasping \cite{King2009}, and suturing \cite{Talasaz2017}. 

Various approaches have been explored to integrate different types of haptic feedback (Figure \ref{fig:feedback_types}) into RMIS. Kinesthetic force feedback has been shown to be effective in certain robot-assisted surgical tasks, such as for palpation \cite{perri2010initial} and suturing \cite{madhani1998black} \cite{wagner2007force}. However, this type of feedback in bilateral teleoperation is susceptible to closed-loop instability due to nonlinear dynamics, friction, among other factors. Cutaneous force feedback systems, such as those that relay interaction forces to the surgeon’s fingertips via skin deformation-based or vibrotactile interfaces, have shown promise in improving task performance and reducing cognitive load. For example, Quek et al. developed a skin deformation tactile feedback device that provided tangential and normal deformations to multiple fingerpads, improving task performance in teleoperated surgical scenarios \cite{Quek2019}. Similarly, Prattichizzo et al. introduced fingertip devices delivering cutaneous feedback, which improved performance in teleoperated tasks by reducing excessive interaction forces and task completion times \cite{Pacchierotti2016}. However, these systems often interfere with the natural use of manipulanda, limiting their adoption in clinical practice. Vibrotactile feedback methods, such as that of VerroTouch, offer stability advantages but may lack the intuitiveness of direct force feedback \cite{Kuchenbecker2010}.

Wrist-worn systems relocate feedback away from the fingers while preserving direct contact between the users' skin and manipulated objects \cite{Palmer2024}. Despite the lower density of mechanoreceptors on the glabrous skin of the forearm as compared to the non-glabrous skin of the fingers, haptic feedback such as localized pressure on the forearm can effectively communicate interactions that would normally be localized to the fingers. Previous work has shown that relocated haptic feedback can assist with perception of mechanical properties of objects in virtual environment \cite{Sarac2022} \cite{DeTinguy2018}. 

In the RMIS setting, these devices could allow surgeons to experience feedback from surgical instruments without encumbering their hands, thereby maintaining the seamless operation of robotic manipulanda. Preliminary studies have demonstrated that such devices can enhance task performance, particularly in virtual and teleoperated environments, by providing haptic cues that improve force modulation and accuracy. For example, Machaca et al. demonstrated that wrist-squeezing force feedback significantly improved accuracy and speed during robotic surgery training tasks, showcasing the potential for skill acquisition and improved task efficiency in RMIS \cite{Machaca2022}.

Despite the potential of wrist-worn haptic systems, significant gaps remain in understanding their efficacy during RMIS tasks. Existing studies primarily focus on virtual environments or generalized teleoperation, with limited exploration of these systems under realistic surgical conditions. Key questions remain regarding the ability of wrist-worn devices to convey accurate and intuitive haptic information, especially when feedback is not co-located with the site of manipulation. Additionally, the impact of such systems on metrics like force error, speed-accuracy tradeoff, and user experience requires further investigation.

This study addresses these gaps by evaluating the effects of a wrist-worn haptic device on force perception and accuracy during teleoperated tasks in RMIS. Participants palpated phantom tissues with the da Vinci Research Kit (dVRK) while wearing a soft pneumatic haptic device on the forearm with an active anchoring bracelet to maintain consistent rendered forces. Performance metrics, including force error and task completion time, were analyzed under conditions with and without haptic feedback. By relocating haptic sensations to the wrist, this study aims to determine whether the device enhances the surgeon’s ability to modulate force and improves overall performance of the task. More specifically, we investigate the effects of wrist-worn haptic feedback on force accuracy and on task speed in teleoperation.



\section{Wearable Haptic Device: Hoxels with Anchoring Bracelet }

\begin{figure}[t]
    \centering
    \begin{minipage}{0.48\textwidth}
        \includegraphics[width=\textwidth]{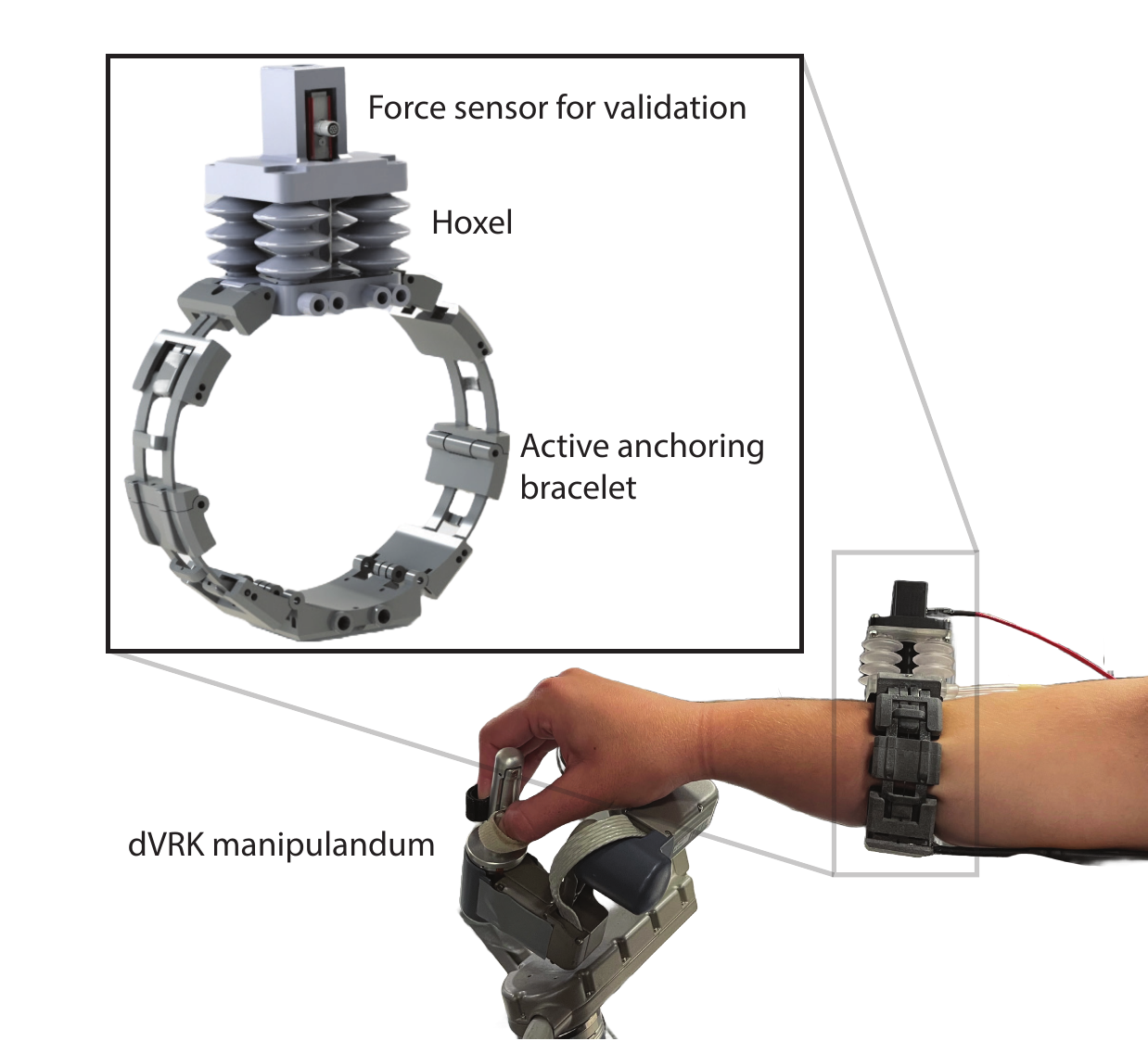}
    \end{minipage}
    \caption{Hoxels device integrated with the active anchoring bracelet. A FUTEK LSB205 S-beam load cell is affixed to an enclosure on the top rigid plate to measure forces exerted by the rigid tactor against the forearm.}
    \label{fig:Hoxel}
\end{figure}

For relocated haptic feedback, we needed a device that could relay significant skin deformation forces on the forearm of the user. We chose to use an iterated design of the haptic voxel displays (Hoxels), first developed by Zhakypov et al. \cite{zhakypov2022hoxels}. In blocked force testing, the device tactor is capable of delivering up to approximately 20 N in normal force against a rigid surface. This range of forces accomplishes more than simple skin deformation, but also applies deep pressure on the forearm that is sufficient to convey haptic forces corresponding to the forces of the teleoperated palpation task. In addition to the Hoxels, we incorporated an active anchoring bracelet to maintain consistent forces from haptic feedback.

\subsection{Hoxel}
\subsubsection{Device Design}

The haptic wearable device in this study is an iterative design based on Hoxels. The device is completely 3D-printed using a commercial stereolithography 3D printer (Form 3, Formlabs Inc., Somerville, MA, USA). It consists of four soft bellow actuators printed with soft material resin (Flexible 80A, Formlabs Inc., Somerville, MA, USA), a rigid bottom plate, rigid top plate with and enclosure for a load cell, and a rigid tactor that is directly screwed onto the load cell (Figure~\ref{fig:Hoxel}). Upon actuation of the bellows via negative pressure, the rigid tactor pushes into the skin of the user's forearm. We improved on the original design of the Hoxels by increasing the size of the bellow actuators to allow for greater vacuum actuation and larger displacement of the tactor. The incorporation of anchoring bracelet with a constant anchoring force prevents the loss of tactor force due to the Hoxels being lifted off the skin surface of the wrist, which was an issue when flexible wristbands were used in early iterations of the device. 

Pneumatic actuation of the Hoxels device was achieved with negative pressure from a building compressor and modulated by a VPPI proportional-pressure regulator (Festo SE \& Co. KG, Esslingen am Neckar, Germany). A Teensy 3.1 microcontroller provides analog input to the Festo VPPI to vary the magnitude of negative pressure to achieve different forces.

\begin{figure}[t]
    \centering
    \vspace{0.4 cm}
    \includegraphics[width=0.48\textwidth]{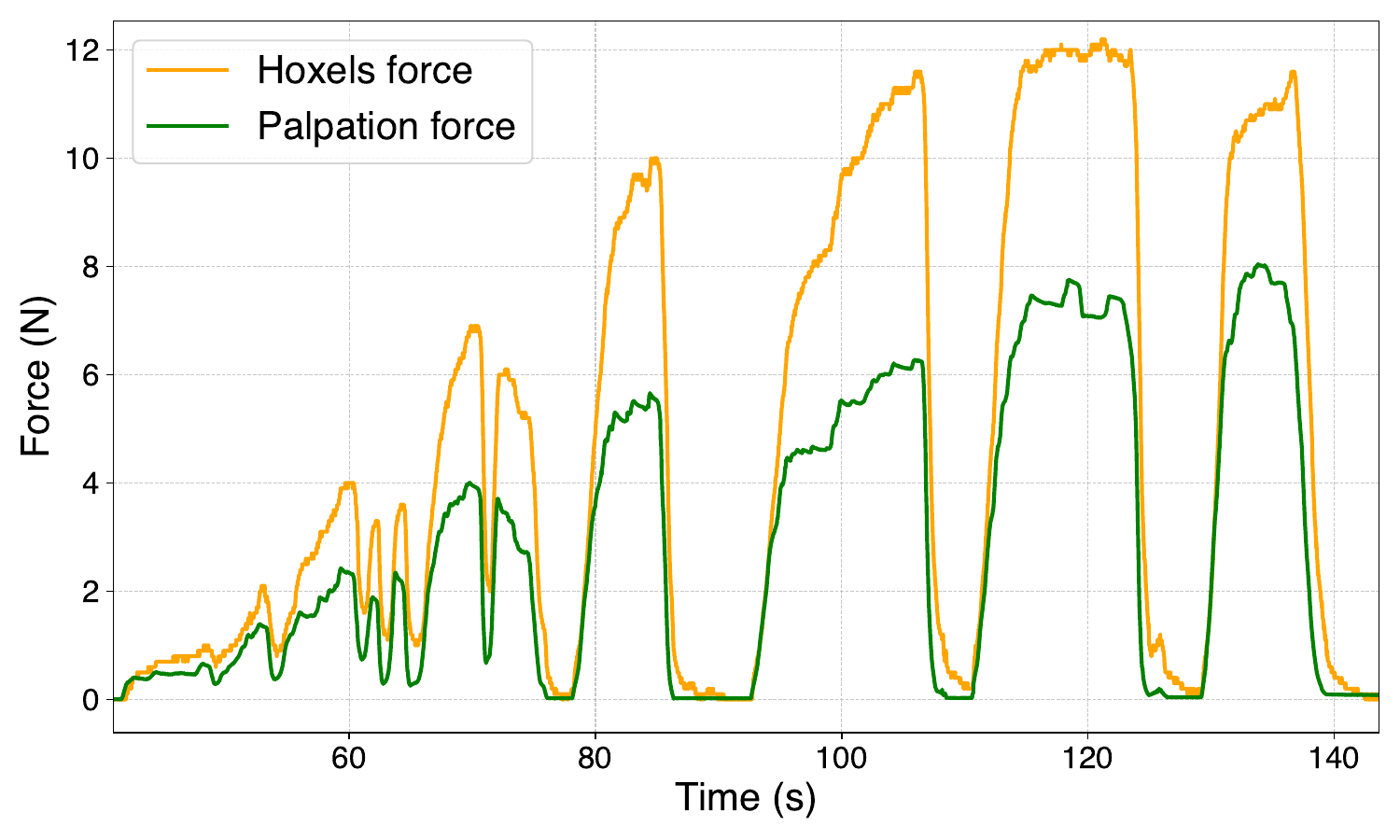}
    \caption{Example of Hoxel haptic feedback forces (as measured by the FUTEK LSB205 load cell) overlaid on the corresponding palpation forces applied on sponge-based phantom tissue (as measured by an ATI Nano17).}
    \label{fig:hoxel_force_profile}
\end{figure}

In order to measure rendered forces in real-time while the Hoxels is worn on the human wrist, we modified the top rigid plate of the Hoxels to incorporate a LSB205 (FUTEK, Irvine, CA, USA) S-beam load cell attached to the sensor enclosure on one end and tactor on the other end. With the addition of the load cell, normal force data was collected while performing a sweepthrough from 0 to 100\% maximum negative presssure, roughly corresponding with 0 to -48.3 kPa negative pressure, with the Festo VPPI, we were able to achieve a maximum of approximately 13.1 N of normal force of the tactor on the dorsal side of the forearm. The mapping derived from this sweepthrough was then used to calculate analog signal commands for the Festo VPPI from desired haptic feedback force levels. We then used this piece-wise linear regression between analog signals and applied forces on the forearm to achieve a linearly scaled mapping between dVRK tool-tissue palpation forces and Hoxels haptic feedback forces rendered on the forearm. This linearly scaled mapping works such that for a palpation force of 7 N, the Hoxels haptically render the maximum of 13.1 N of normal force on the forearm. For higher palpation forces, the user only receives the maximum of 13.1 N. An example of Hoxel haptic feedback forces experienced on the forearm of a user during a series of palpation movements is shown in Figure~\ref{fig:hoxel_force_profile}.

\subsection{Active Anchoring Bracelet}
\subsubsection{Bracelet Design}
\begin{figure}[t]
    \centering
    \vspace{0.4 cm}
    \includegraphics[width=0.45\textwidth]{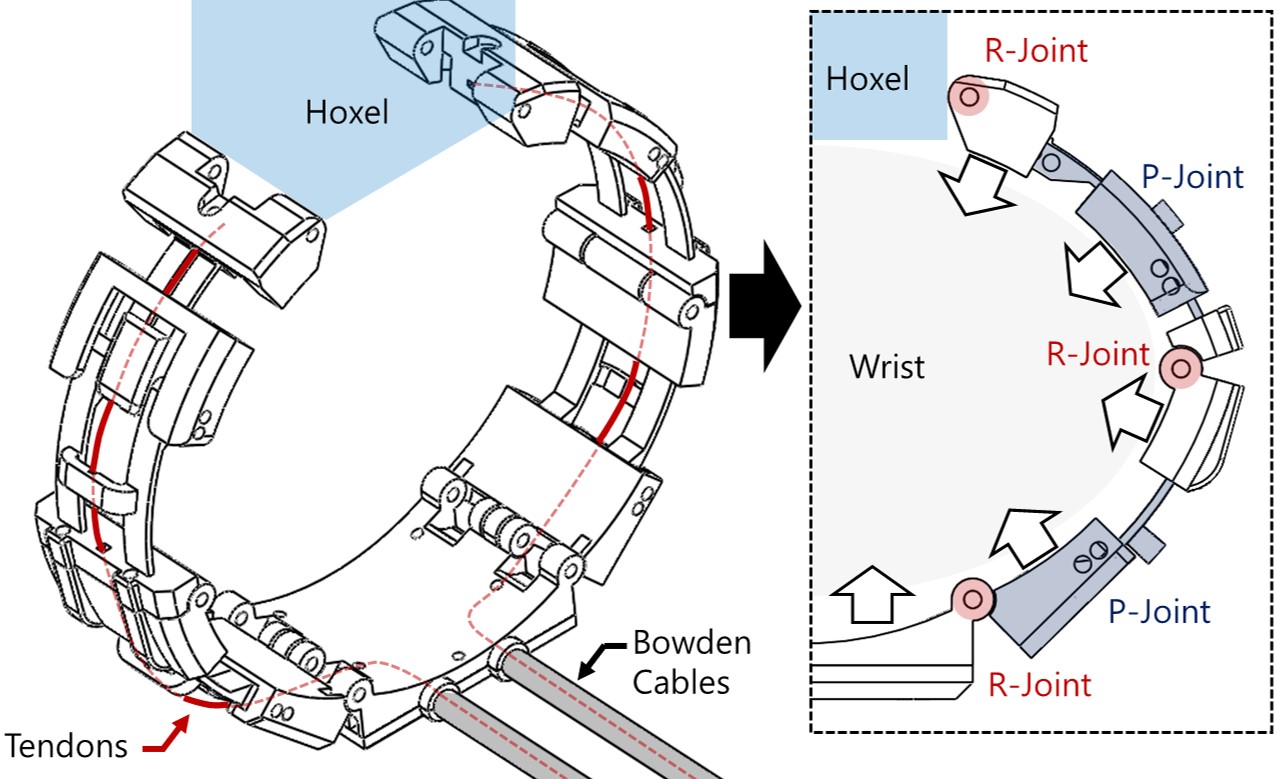}
    \caption{Overview of the anchoring bracelet structure. The bracelet is underactuated by two tendons, allowing it to securely fasten the Hoxels to the user's forearm while remaining shape-adaptive.}
    \label{fig:anchor_mechanism}
\end{figure}

The active anchoring bracelet (Figure~\ref{fig:anchor_mechanism}) consists of two wings connected to a base structure, with the wing tips holding the Hoxel. Each wing features three revolute joints and two prismatic joints and is underactuated by a single tendon. The base structure connects the bracelet to the Bowden cables, which transmit the tendons to the actuator. Within the base structure, the tendons follow a smooth 90-degree bend. The prismatic joints in the wings allow the bracelet to expand when the user is donning the device and contract when fastening it to the forearm. The revolute joints enable shape adaptation of the bracelet and help address geometric constraints related to prismatic joint displacements. To improve wearability, we added neoprene sponges inside the bracelet for cushioning.

\subsubsection{Tendon Actuator}
A slack-enabling actuator based on the work of Cheon et al.\cite{cheon2020single}, was used to prevent tendon derailment from the spool when the tendons were not under tension. The actuator employs a silicone-based jamming mechanism to feed the tendons while securing them in place to avoid derailment. Additionally, this feature enables the actuator to apply a pushing force to the anchoring bracelet during release. To efficiently transmit this force to the bracelet structure, we used a stiff, single-stranded nitinol wire as the tendon. This releasing action allows the user to easily don and doff the bracelet.

\section{Study Design and Procedure}

In order to evaluate the effects of wrist-worn relocated haptic feedback on force accuracy and task speed in a teloperated manipulation task, we conducted a user study involving a simple palpation task with the dVRK while users wore the Hoxels with active anchoring bracelet. The study consisted of a learning phase, during which participants were acquainted with the haptic feedback and learned to perform the teleoperated palpation task, and an evaluation phase. A total of 24 participants, between the ages of 19 and 62, participated in the study. All participants were right-handed and were novices in that they had no prior experience teleoperating the da Vinci Research Kit or da Vinci Surgical System (Intuitive Surgical, Inc., Sunnyvale, CA, USA).
\subsection{Setup}
\subsubsection{Hardware}
Participants used the first-generation (Classic) da Vinci Research Kit (dVRK) to perform a teleoperated palpation task. The surgeon console consists of two 8-DOF master tool manipulators (MTMs), a foot-pedal tray, and a CRT-monitor stereoviewer (640x480 resolution). The patient side consists of two 7-DOF patient-side manipulators (PSMs) and a ZED Mini camera (StereoLabs, San Francisco, CA, USA) to provide the operator with a 3D stereoscopic view of the workspace. For the palpation task in this study, participants only used the right MTM (MTMR) to teleoperate the right PSM (PSM1) with EndoWrist ProGrasp Forceps (Intuitive Surgical, Inc., Sunnyvale, CA, USA) attached as the end-effector tool. 

For the phantom tissue, a sponge-based skin pad (Limbs \& Things, Bristol, UK) was fixed in a 3D-printed enclosure with a 6-axis Nano17 force-torque sensor (ATI Industrial Automation, Apex, NC, USA) affixed to the center of the enclosure's bottom surface. The phantom tissue-sensor enclosure was securely attached to an 80-20 mount placed in the center of the workspace.

\begin{figure}[t]
    \centering
    \vspace{0.4 cm}
    \includegraphics[width=0.48\textwidth]{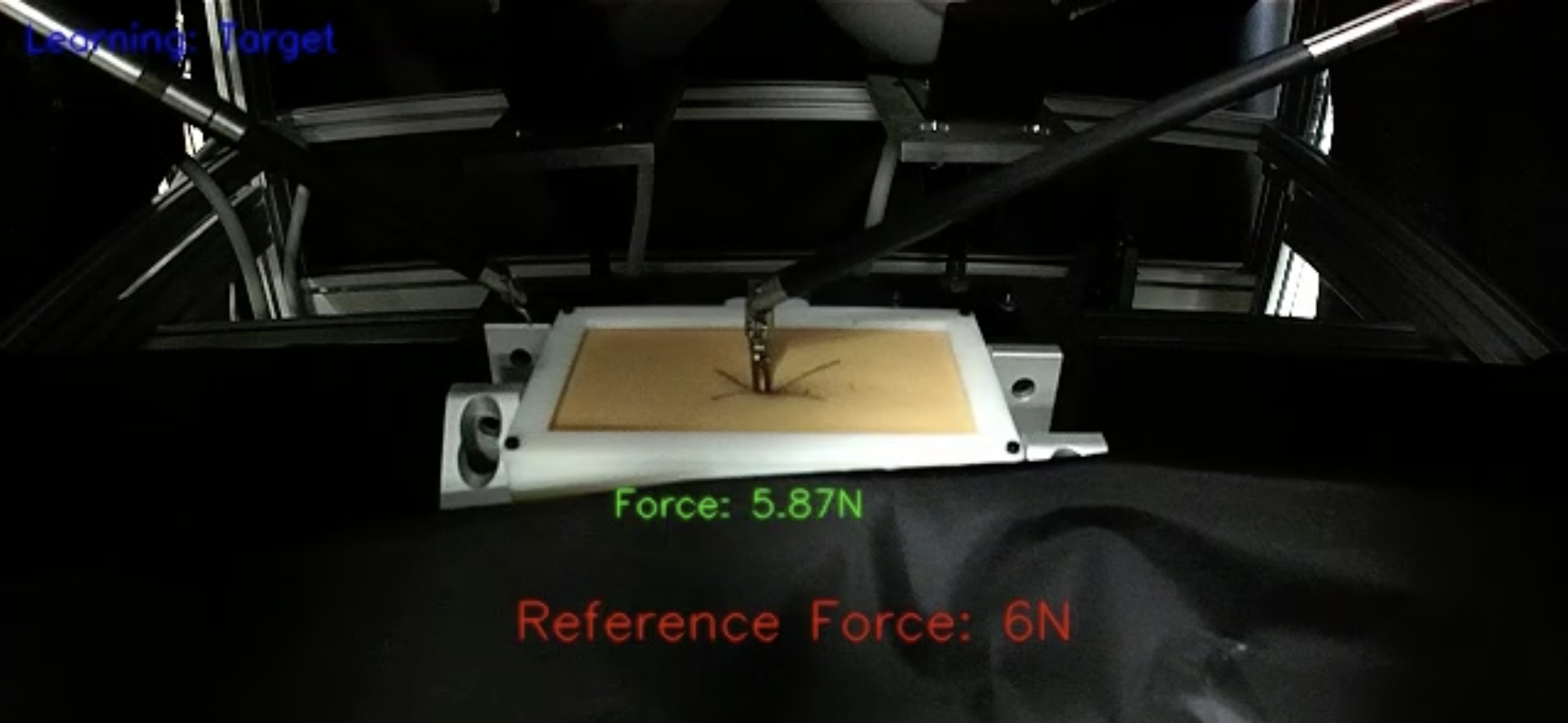}
    \caption{Study participant's stereoscopic camera view of the patient side of the dVRK through the visual console. The phantom tissue-sensor enclosure is placed in the center of the workspace. During the experiment, participants palpate the center of the tissue pad with the tip of the ProGrasp forceps attached to the PSM.}
    \label{fig:patient_side}
\end{figure}

Prior to the participant beginning each part of the experiment, the position of the PSM was initialized so that the forceps tip would be directly above the center of the phantom tissue. To interact with the GUI on the stereoviewer, participants use a keyboard affixed to the arm rest of the surgeon console.

\subsubsection{Software}
For dVRK teleoperation, a ROS-based open controller implementing proportional-derivative position control \cite{chen2017software}. A scaling factor of 0.2 was used to scale motions from the MTMR to PSM1.

A custom Python script ran on a PC to process force data received by the Nano17 force-torque sensor, perform the haptic rendering calculations, and send commands over serial to a Teensy 3.1 for pneumatic actuation of the Hoxel. The same script generated a graphical interface projected onto the surgeon console stereoviewer (Figure~\ref{fig:patient_side}). This graphical interface provided the participants with real-time displays of the palpation forces and textual cues overlaid on the stereoscopic camera view of the workspace. 

During the study, using a CAN communication protocol-based interface, the experimenter sent commands for tightening and release of the anchor, monitored tendon transmission position and motor torque in real time, and monitored the forces of the Hoxels haptic feedback on the wrist.

\subsection{Protocol}
This study consisted of learning and evaluation phases. Participants spend a maximum of one hour completing the entire experimental procedure. The university Institutional Review Board approved the experimental protocol and all participants gave informed consent.

\subsubsection{Learning}
To ensure that participants had adequate practice with the teleoperated palpation task while experiencing relocated haptic feedback on the forearm, participants first went through a learning phase. The four parts of this learning phase involved different combinations of the presence of relocated haptic feedback and visual display of the applied force numerical value. The purpose of the ``Passive Haptics" phase was to acquaint participants with the magnitudes of relocated feedback associated with the magnitudes of applied force in the palpation task. The second part, ``Explore" phase, was intended to allow participants to associate relocated haptic feedback on the forearm with tool-tissue interactions in the teleoperated palpation task. Participants then continued to learn the force-matching palpation task in ``Target" phase while still receiving visual feedback with numerical display of forces before finally practicing the task in a ``Practice" phase as it would appear in the evaluation phase.
\paragraph{Passive Haptics} Participants passively experienced wrist-worn haptic feedback corresponding to five reference force levels of tissue palpation: 2N, 3N, 4N, 5N, and 6N. Each reference force level was repeated three times, in a random order. The participant could spend as much time as needed to experience the haptic feedback corresponding to each force level and would proceed to the next force level by hitting the ENTER key on a keyboard.
\paragraph{Explore} Participants learned how to teleoperate the dVRK and perform the palpation task. They were given 2 minutes to explore palpation while experiencing wrist-worn haptic feedback and seeing the real-time palpation force numerical display on screen. Participants were instructed to palpate in a manner to experience the full range of haptic feedback but avoid generating palpation forces above 8 N. Participants were also instructed to move the tool only along the vertical axis above the center of the tissue pad to ensure forces were applied normal to the tissue surface.
\paragraph{Target} Participants next learned to perform the force-matching palpation task. Participants were instructed to palpate to match the reference force levels displayed on screen, each repeated three times in a random order. They received a real-time numerical display of the palpation force as well as wrist-worn haptic feedback. The participant would indicate they have matched the force by hitting the ENTER key on a keyboard.
\paragraph{Practice} In this phase, participants practiced the force-matching palpation task similar to how it would be performed in the evaluation phase. While participants received real-time wrist-worn haptic feedback during the palpation, unlike the Target phase, they did not receive a real-time numerical display of the palpation force. Instead, the actual palpation force would only be displayed once the participant has indicated they have settled on a final palpation force.
\subsubsection{Evaluation}
For the evaluation phase, participants performed the force-matching palpation task either with (``Haptics", H) or without haptics (``No Haptics", NH). No real-time numerical display of palpation forces would be provided at any point in this part of the experiment. The order of the conditions (H, NH) was randomized for each participant. Of the 24 participants in the study, 12 completed the H condition first, the other 12 completed the NH condition first.
\subsection{Metrics}
\paragraph{Force Accuracy} To evaluate force accuracy in the palpation task, the primary metric analyzed was force error, defined as the magnitude of difference between the reference force level and actual palpation force at the moment participants indicated they have matched the desired force. The average force error was calculated across all subjects for each condition (H, NH) per reference force level (2N, 3N, 4N, 5N, 6N) and aggregate across all reference force levels. 
\paragraph{Task Speed}

\begin{figure}[t]
    \centering
    \hspace*{-0.04\textwidth}  
    \includegraphics[width=0.55\textwidth]{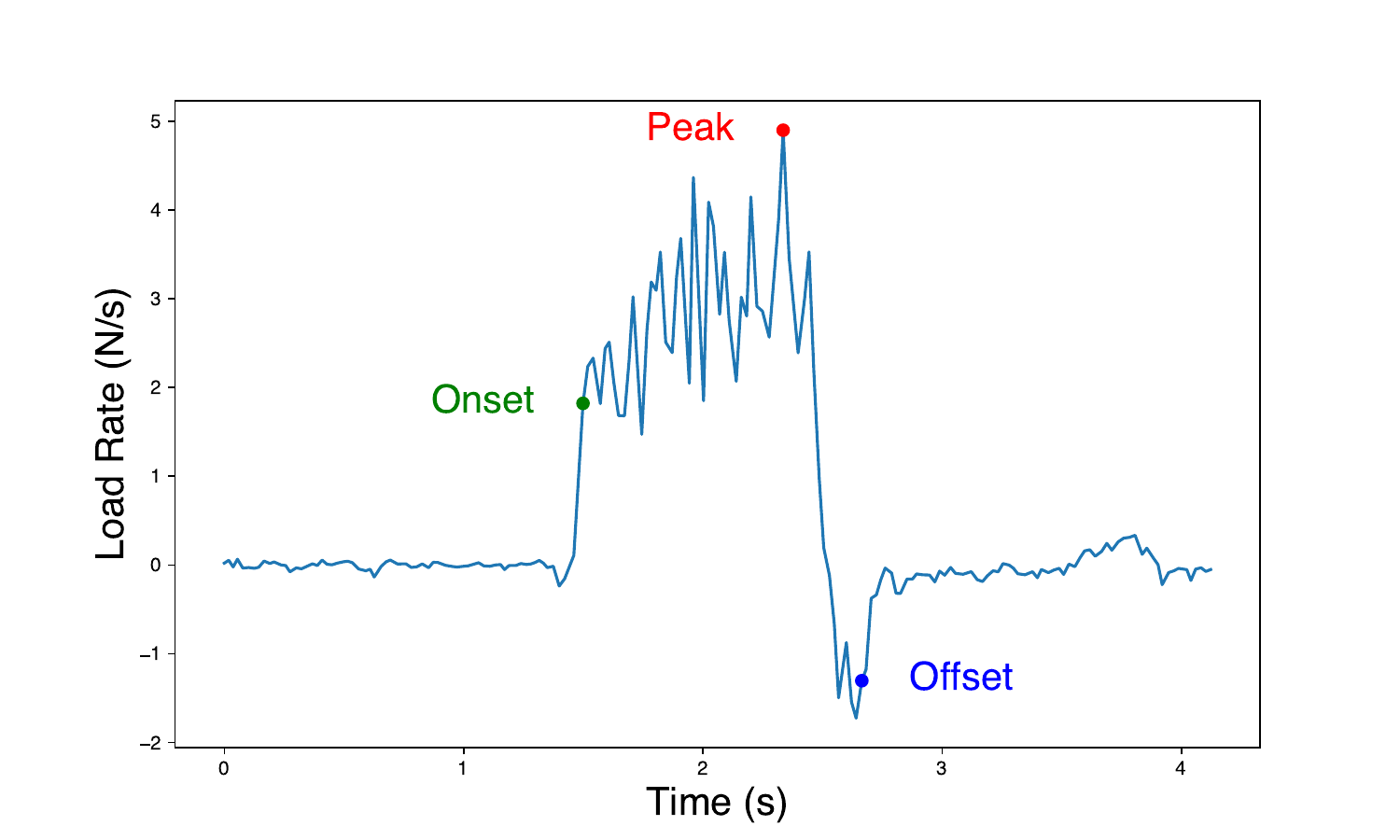}
    \caption{Load rate (LR) over time during a palpation for 6N. Onset was when load rate first crossed 25\% of the peak LR while offset was when load rate last crossed 25\% of the peak LR. Trial time (2.02 s for this example) was calculated as the difference between offset and onset times.}
    \label{fig:lr_profile}
\end{figure}

The speed of the palpation task was measured using the movement time and load rate. Load rate is calculated as the change in palpation force over time, with the peak load rate of a single palpation being the maximum value. In the palpation task, movement onset and offset were computed by identifying the times when the load rate first and last, respectively, crossed $\pm 25\%$ of the peak load rate, as shown in Figure~\ref{fig:lr_profile}. According to Fitts's law \cite{fitts1954information}, slower movements are often more accurate. In order to characterize the speed-accuracy tradeoff, we use a normalized metric, the normalized absolute error-time (NAET):
\begin{equation*}
\text{NAET} = \frac{\lvert \text{error} \rvert}{\lvert \text{error}_{\text{max}} \rvert} \times \frac{\text{time}}{\text{time}_{\text{max}}}
\end{equation*}
where $\text{error}_{\text{max}}$ and $\text{time}_{\text{max}}$ are the observed maxima for each quantity during evaluation of across all subjects within a condition (H, NH) for each reference force level (2N, 3N, 4N, 5N, or 6N). The normalization of force error magnitudes and movement times by their respective maxima produces unitless values ranging from 0 to 1, ensuring equal importance is assigned to both accuracy and time. The closer the NAET value is to 0, the less of a trade-off there is between speed and accuracy compared to an NAET value closer to 1.

\subsection{Statistical Analysis}
For each performance metric, we fit an ordinary least squares regression with fixed effects of the haptic condition (H, NH), experimental order of the haptic conditions during evaluation (H first, NH first), along with the interactions of these effects. 

Metrics including force error, palpation movement time, and normalized accuracy-error time (NAET) were analyzed for statistical signficance. An analysis of variance (ANOVA) was conducted for each model using Type II sum of squares to evaluate the significance of fixed effects. Analyses were conducted separately for each reference force level as well as for the aggregated data, with significance defined as p-value \ensuremath{<} 0.05.
 
\section{Results and Discussion}

\subsection{Effects of Wrist-Worn Haptic Feedback on Force Accuracy}

\begin{figure}[t]
    \centering
    \vspace{0.6 cm}
    \includegraphics[width=0.5\textwidth]{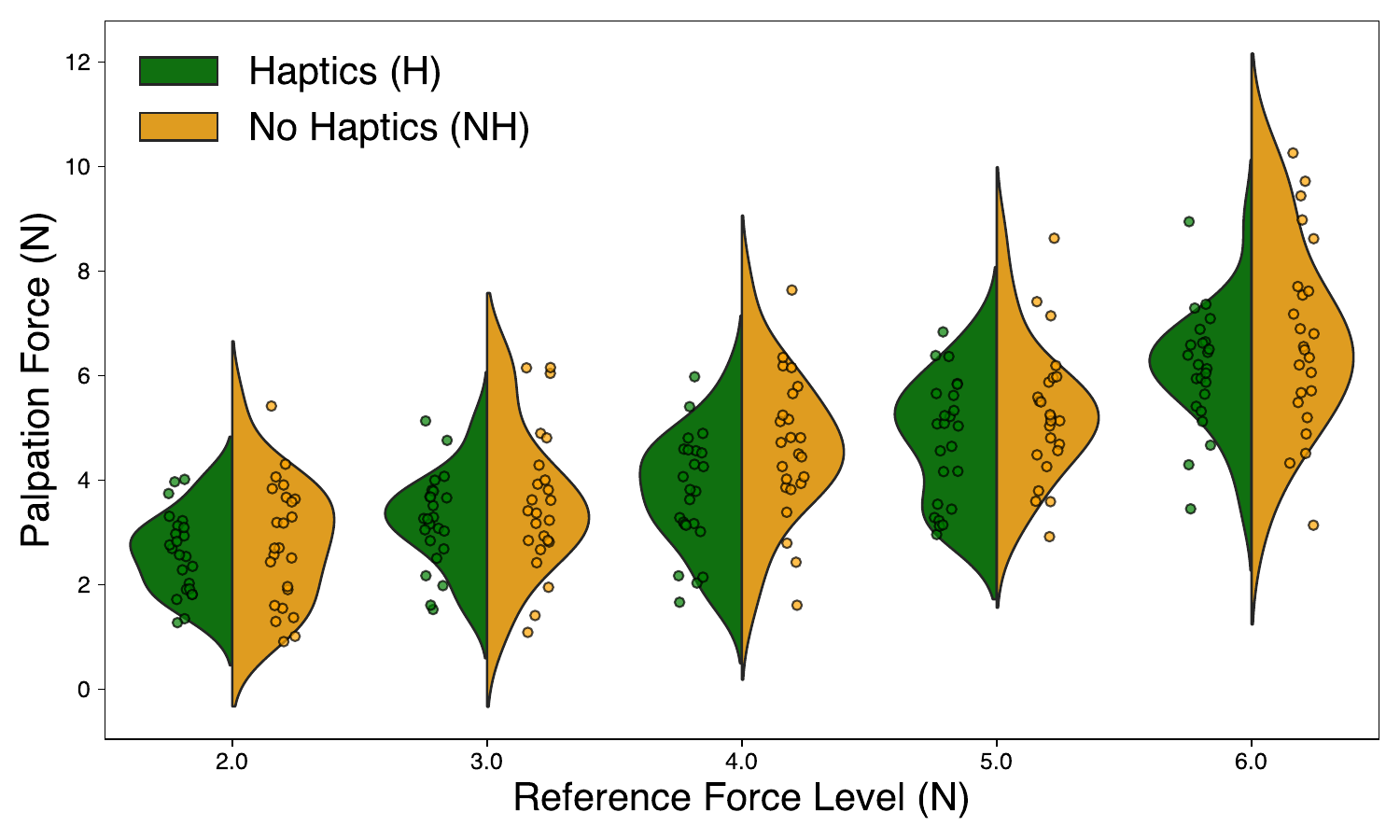}
    \caption{Participants consistently applied higher forces when performing the palpation task without haptics (NH) as compared to with haptic feedback (H).}
    \label{fig:violin_plots}
\end{figure}

Based on the distribution of palpation force across all reference force levels for all subjects (Figure~\ref{fig:violin_plots}), participants consistently applied higher palpation forces without haptic feedback (NH) as compared to with haptic feedback (H). The maximum force applied by participants in the NH condition was 10.75N compared to only 8.40 N in the H condition, and both maxima occurred during a palpation for reference force level 6N. Without haptic feedback (NH), the palpation forces have a wider distribution, indicating a larger variability in the forces applied by participants. Palpation forces show a more concentrated distribution in the H condition, most notably at the force levels of 2N, 3N, and 6N, indicating a more consistent application of force when wrist-worn haptic feedback is provided. As the reference force levels increased, the difference in the distribution of the forces between conditions remained relatively constant. 

\begin{figure}[t]
    \centering
    \includegraphics[width=0.5\textwidth]{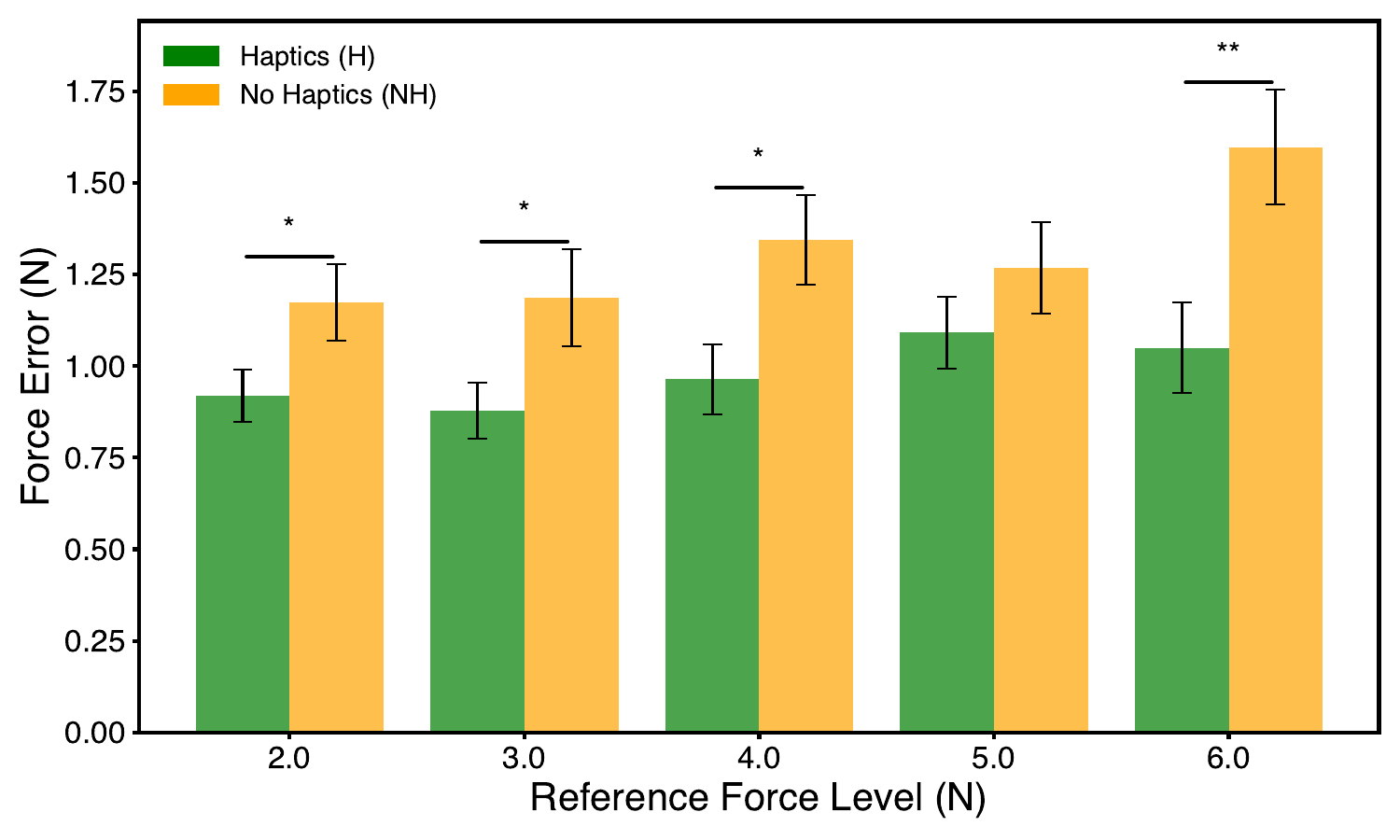}
    \caption{Mean force errors in evaluation phase. At all reference force levels except 5N, participants had a lower force error when performing the task with haptics (H) as compared to without haptics (NH).}
    \label{fig:force_error}
\end{figure}

In terms of magnitude of palpation force error, participants had a smaller force error when performing the palpation task in the H condition compared to the NH condition (Figure~\ref{fig:force_error}). There was a significant effect of haptic condition on force error at the reference force levels of 2N (\textit{p} \ensuremath{<} 0.05), 3N (\textit{p} \ensuremath{<} 0.05), 4N (\textit{p} \ensuremath{<} 0.05), 6N (\textit{p} \ensuremath{<} 0.01), and the aggregate force error across all force levels (\textit{p} \ensuremath{<} 0.001). The difference in force error was most pronounced at reference force level of 6N, where force error averaged at approximately 1.60 N for the NH haptics condition but only 1.00 N for the H condition. 

The overall lower palpation forces in the H condition suggest that wrist-worn haptic feedback not only improves the accuracy of force perception but also has an inhibitive effect on application of forces higher than the target force. Without haptic feedback, participants likely overestimate the required force due to reliance on visual and proprioceptive cues, leading to higher force application. The reduction in applied force is consistent with previous works that demonstrate the efficacy of tactile feedback in reducing grip force \cite{King2009} \cite{King2009TOH}, contact forces \cite{khursid2017}, and normal force \cite{wagner2002} with telerobotic and telesurgical tasks. The reduction of applied has implications on the capability to avoid excessive forces when handling delicate tissue in telerobotic surgery. 

The reduced force error with wrist-worn haptic feedback aligns with findings in previous studies that demonstrate the efficacy of wrist-based haptic feedback in improving perception of force differences in both virtual environments \cite{pezent2019} \cite{Sarac2022} \cite{ercan2025wrist} and teleoperation settings \cite{brown2017} \cite{Machaca2022}. Without wrist-worn haptic feedback, participants mostly relied on visual and proprioceptive cues in the NH condition.  Visual and proprioceptive cues are known to be less reliable for fine force discrimination, especially in the absence of co-located tactile input \cite{kuschel2010combination} \cite{Klatzky2014}. 

The wider distribution in palpation forces in the NH condition may reflect the increased cognitive demand of force estimation without real-time haptic feedback, thus resulting in more difficulty with maintaining consistent forces. Previous work has shown that sensory substitution can improve the consistency of robotically applied forces in suture-tying with the da Vinci system \cite{kitagawa2005effect}. This consistency of applied forces is an indicator of better control by surgeons over surgical tasks.

Prior studies have shown that, when deprived of haptic feedback, users exhibit inconsistent, often excessive, forces due to the difficulty of gauging and reproducing precise force levels \cite{King2009} \cite{perri2010initial} \cite{Machaca2022}. King et al. \cite{King2009} found that surgeons applying forces without tactile feedback exerted higher grasping forces, while introducing a tactile feedback system reduced and stabilized their grasping force. Machaca et al. \cite{Machaca2022} showed that wrist-squeezing force feedback helped novices regulate their force output and consistently apply lower forces during robotic surgical training. These findings suggest that the presence of wrist-worn haptics anchors user perception and modulates motor output, enabling not only more accurate but also more consistent force application \cite{bergholz2023benefits}.

\subsection{Effect of Haptics on Task Speed}

\begin{figure}[t]
    \centering
    \includegraphics[width=0.5\textwidth]{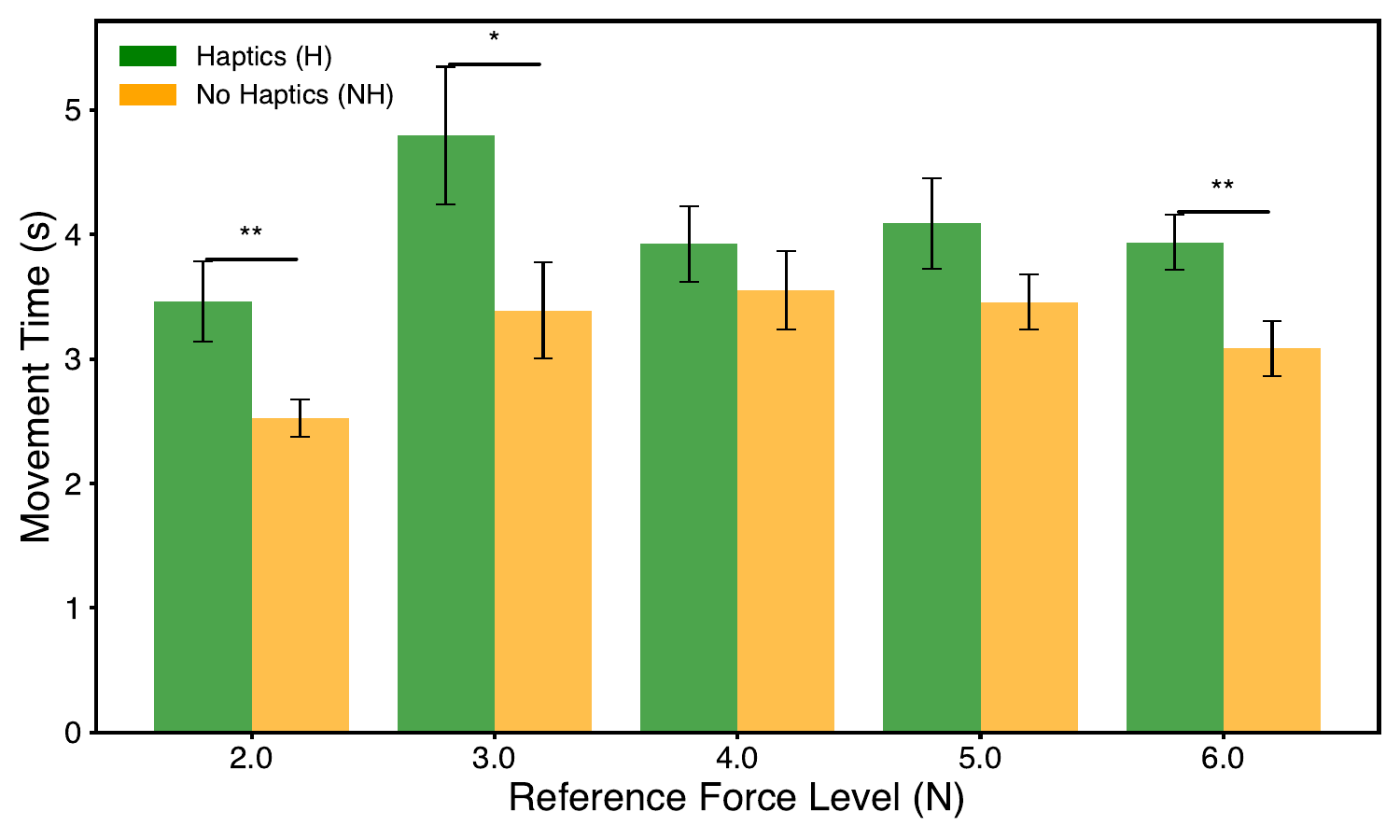}
    \caption{Mean movement times in evaluation phase. Movement time is higher with haptics (H) than without haptics (NH) at 2N, 3N, and 6N reference force levels.}
    \label{fig:movement_time}
\end{figure}

There was a significant effect of haptic condition on palpation movement time at the reference force levels of 2N (\textit{p} \ensuremath{<} 0.01), 3N (\textit{p} \ensuremath{<} 0.05), 6N (\textit{p} \ensuremath{<} 0.01), and the aggregate palpation time across all force levels (\textit{p} \ensuremath{<} 0.001) (Figure~\ref{fig:movement_time}). With wrist-worn haptic feedback (H), movement times were generally longer than without haptic feedback (NH).
    
\begin{figure}[b]
    \centering
    \includegraphics[width=0.5\textwidth]{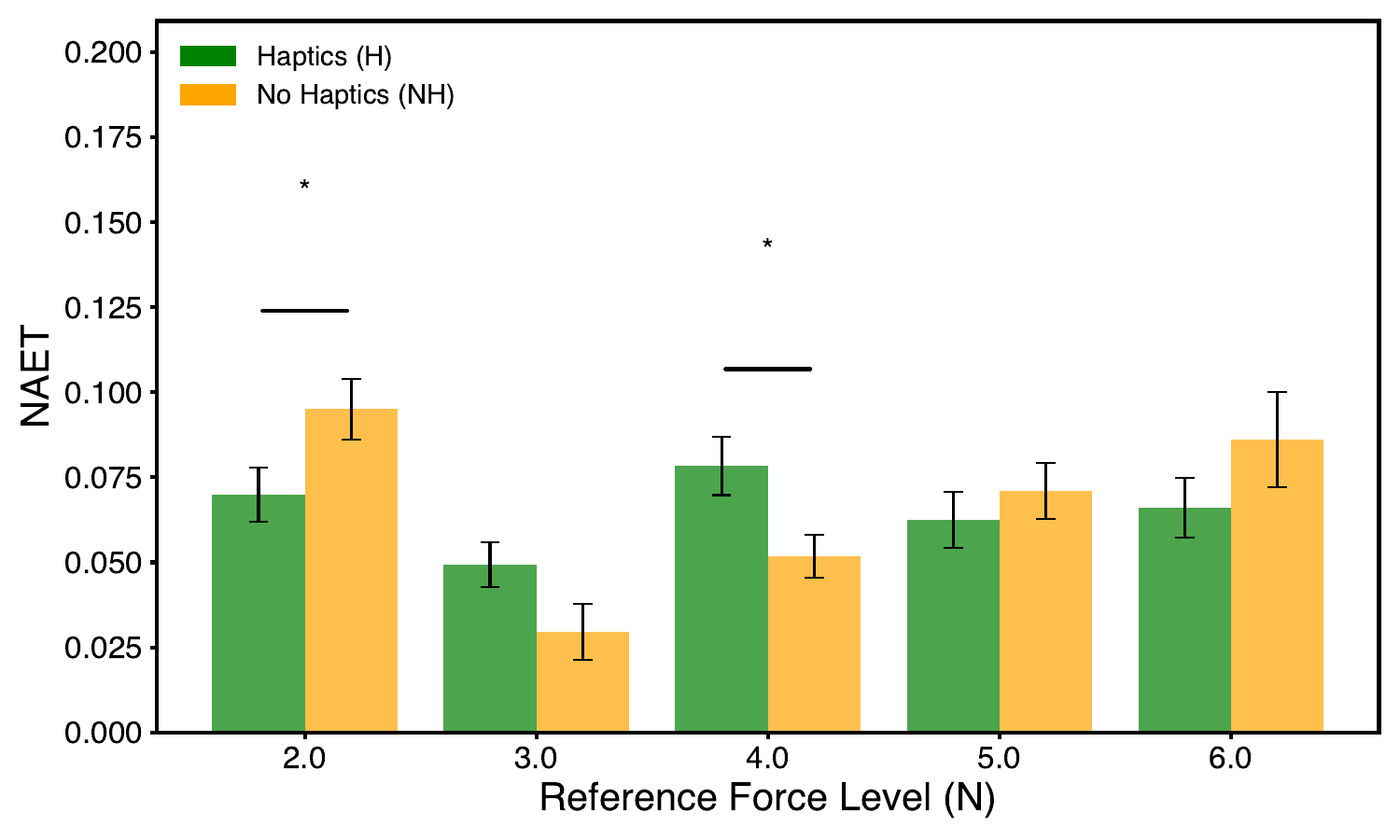}
    \caption{Mean NAET in evaluation phase. There are only statistically significant differences in normalized accuracy-error time (NAET) between haptics (H) and no haptics (NH) conditions at references force levels 2N and 4N.}
    \label{fig:NAET}
\end{figure}

Inspecting the normalized accuracy-error time (NAET), there were no salient trends in terms of differences between the haptic conditions. There was significant effect of wrist-worn haptic feedback on NAET at only reference force levels 2N (\textit{p} \ensuremath{<} 0.05) and 4N (\textit{p} \ensuremath{<} 0.05) (Figure~\ref{fig:NAET}). However, these significant differences showed opposing phenomena, as the speed-accuracy tradeoff was less for H condition at 2N but was less for the NH condition at 4N. The results indicate there was more of a speed-accuracy tradeoff with haptic feedback when palpating to the 4N reference level as compared to without haptic feedback, whereas the opposite was observed when palpating to 2N.

Longer movement times in the haptics condition suggest that participants were more deliberate when modulating their forces. Participants may take more time to confirm force levels when receiving wrist-worn haptic feedback compared to purely relying on visual estimation and proprioceptive cues. The increase in movement time is consistent with the effect of haptics on task completion time in other teleoperation studies. Ortmaier et al. showed that providing force feedback during a robotic dissection reduced unintended injuries but increased the operation time compared to manual dissection \cite{ortmaier2007robot}.

The lack of a consistent trend across force levels indicates that participants dynamically adjusted their control strategies depending on the availability of feedback and the difficulty of the force target. According to Fitts's Law \cite{fitts1954information}, participants typically achieve higher accuracy with reduced movement speed.  Significant differences in NAET at 2N and 4N but not at other levels suggest that the influence of haptic feedback on the speed-accuracy tradeoff is potentially dependent on the force level. Previous research has shown that the benefit of haptic feedback can depend on task difficulty and user skill: more challenging or higher-force tasks potentially induce larger slow-downs in movement to improve accuracy \cite{bergholz2023benefits}. Overall, with haptic feedback, participants achieved higher accuracy at lower movement speeds. 

\section{Conclusions}

This study tested the effects of relocated haptic feedback on the accuracy of applied force and speed during the teleoperation of a minimally invasive surgical robot in a simple palpation task, compared against a control condition without haptic feedback. Our results pointed to a strong effect of wrist-worn haptic feedback in reducing the overall magnitude of palpation force as well as decreasing error when aiming for target force levels. Additionally, the presence of haptic feedback led to longer movement times, suggesting that participants were more deliberate in their force modulation, thereby improving force application accuracy. While no consistent trend was observed in the normalized accuracy-error time (NAET) across all force levels, significant differences at select force levels suggest that the influence of haptic feedback on task efficiency may depend on task difficulty or sensitivity to force discrimination.

Future work should verify our findings in more complex manipulation tasks that involve force application in multiple degrees of freedom (DOF), such as tool guidance, blunt dissection, or suture handling. In addition to the 1-DOF normal force rendering employed in this study, future studies should explore the effects of multi-DOF relocated haptic feedback, particularly the inclusion of shear and tangential forces, on performance and perception in teleoperation. Expanding the design and evaluation of relocated haptics across diverse tasks and force directions will be critical for assessing their viability as effective alternatives to fingertip-based feedback in robot-assisted surgery.

The findings of this study contribute to the growing body of research on haptic feedback in RMIS and offer insights into the design and integration of wearable haptic systems. By demonstrating the feasibility and benefits of wrist-worn feedback, this work lays the foundation for advancing effectiveness of haptic feedback to improve force perception, applied force accuracy, and overall performance in teleoperated manipulation tasks.

\bibliographystyle{IEEEtran}
\bibliography{references}

\end{document}